\documentclass[10pt, conference]{IEEEtran}
\IEEEoverridecommandlockouts
\usepackage{cite}
\usepackage{amsmath,amssymb,amsfonts}
\usepackage{algorithmic}
\usepackage{textcomp}
\usepackage{xcolor} 
\definecolor{LightGray}{gray}{0.9}
\usepackage{todonotes}
\usepackage{lipsum}
\usepackage{float}
\usepackage{flushend}
\usepackage{makecell}

\def\BibTeX{{\rm B\kern-.05em{\sc i\kern-.025em b}\kern-.08em
    T\kern-.1667em\lower.7ex\hbox{E}\kern-.125emX}}

\usepackage{graphicx}
\graphicspath{{.}}

\usepackage[english]{babel}
\usepackage{booktabs}  
\usepackage{multirow}  
\usepackage{amsmath}  
\usepackage{mathtools}

\makeatletter
\def\namedlabel#1#2{\begingroup
    #2%
    \def\@currentlabel{#2}%
    \phantomsection\label{#1}\endgroup
}
\makeatother

\begin{document}

\title{Enhancing Spatiotemporal Networks with xLSTM:\\A Scalar LSTM Approach for Cellular Traffic Forecasting}

\author{\IEEEauthorblockN{Khalid Ali, Zineddine Bettouche, Andreas Kassler, Andreas Fischer}
    \IEEEauthorblockA{Deggendorf Institute of Technology \\
        Dieter-Görlitz-Platz 1, 94469 Deggendorf \\
        \{khalid.ali, zineddine.bettouche, andreas.kassler, andreas.fischer\}@th-deg.de}
}

\maketitle

\begin{abstract}
  Accurate spatiotemporal traffic forecasting is vital for intelligent resource management in 5G and beyond. However, conventional AI approaches often fail to capture the intricate spatial and temporal patterns that exist, due to e.g., the mobility of users. We introduce a lightweight, dual-path Spatiotemporal Network that leverages a Scalar LSTM (sLSTM) for efficient temporal modeling and a three-layer Conv3D module for spatial feature extraction. A fusion layer integrates both streams into a cohesive representation, enabling robust forecasting. Our design improves gradient stability and convergence speed while reducing prediction error. Evaluations on real-world datasets show superior forecast performance over ConvLSTM baselines and strong generalization to unseen regions, making it well-suited for large-scale, next-generation network deployments. Experimental evaluation shows a 23\% MAE reduction over ConvLSTM, with a 30\% improvement in model generalization. 
\end{abstract}


\section{Introduction}\label{sec:intro}
The surge in the number of mobile devices and the advent of data-intensive applications have sharply increased traffic volume and variability in 5G networks. Accurate, fine-grained spatiotemporal traffic forecasting is crucial for effective network management, including demand-aware resource allocation, congestion control, and proactive infrastructure planning. For example, precise forecasts enable operators to dynamically scale computing and radio resources to match demand, thereby reducing operational costs and improving utilization efficiency~\cite{chen2020intelligent, zhao2021predictive}. Moreover, long-term forecasts inform strategic decisions, e.g., base-station deployment and antenna configuration, which rely on a solid understanding of persistent traffic patterns across urban regions~\cite{dong2020millimeterwave, mei2023joint}. Traditional timeseries forecasting methods (e.g., ARIMA and exponential smoothing) have relied on univariate timeseries~\cite{arima}, where a single variable's past values are used to predict its future trajectory. 

Recently, machine learning-based methods, such as Long Short-Term Memory (LSTM), have shown great potential in capturing nonlinear temporal dependencies~\cite{lstm}. By combining LSTM layers with convolutional feature extractors, hybrid methods such as LSTM-FCN and its attention-augmented variant ALSTM-FCN demonstrated gains in accuracy by integrating local pattern detection with sequence modeling~\cite{karim2018lstmfcn}\cite{karim2019multivariatelstmfcn}. Such methods are particularly suited for network traffic forecasting, where both temporal trends and localized patterns matter. Kernel-based methods, e.g., Rocket and MiniRocket~\cite{dempster2020rocket,dempster2021minirocket}, utilize random kernels and simple classifiers to realize improved univariate performance with minimal computational overhead. Despite their successes, univariate approaches inherently ignore spatial context. In mobile networks, traffic in one cell depends on both its history and the neighboring dynamics caused by handovers and mobile users. Empirical studies show that integrating spatial heterogeneity is essential for capturing the true complexity of mobile traffic patterns~\cite{silvestrini2008temporal}. Moreover, approaches that treat each cell in the base station grid as a multivariate channel yield only modest gains because they lack explicit mechanisms for long-range interactions~\cite{ruiz2021great,wen2022transformers}.

To address these limitations, a new class of forecasting techniques emerged that specifically aim to capture both spatial and temporal dependencies. Spatiotemporal architectures—ConvLSTM, STN, graph-based models (e.g., MTGNN), and transformers (e.g., Spacetimeformer)—explicitly model these interactions and outperform univariate adaptations~\cite{ConvLSTM_NIPS15,stn,wu2020connecting,grigsby2021longrange}. Despite their advantages, ConvLSTM-based methods show limited scalability due to their high computational cost and slow convergence. Hence, scalar LSTM (sLSTM), a component of the extended LSTM (xLSTM), is introduced to reduce parameter complexity and computational cost while improving convergence and long-term temporal dependencies modeling~\cite{xlstm}. In this paper, we propose a new STN-based model that integrates sLSTM modules and augments its fusion mechanism. Our key contributions are:
\begin{itemize}
    \item We introduce STN-sLSTM-TF, a novel spatiotemporal model using sLSTM for efficient convergence and attention fusion for long-range dependencies.
    \item We define and compare the model variants, STN-sLSTM, STN-TF, and STN-sLSTM-TF, to isolate component impacts on accuracy and efficiency.
    
    \item We apply our approach to a large-scale cellular traffic dataset (Italia Telecom), demonstrating its effectiveness and generalization across diverse environments, achieving a 23\% reduction in MAE compared with baselines.

\end{itemize}

Together, our enhanced STN achieves improved spatiotemporal performance in the context of network traffic.

\section{Background and Related Work}\label{sec:related-work}
Traditional timeseries forecasting methods like Exponential Smoothing and ARIMA rely on linear timeseries regression; Holt-Winters exponential smoothing was used for short-term forecasting~\cite{Vlahogianni2020}, while ARIMA models were applied to predict base station loads, often outperforming LSTMs in purely temporal settings~\cite{BaseStationARIMA2024}. However, such techniques treat each site independently, overlooking important spatial correlations~\cite{Vlahogianni2020}. Recent surveys have highlighted advanced factor analysis and tensor decomposition methods to uncover complex spatiotemporal structures~\cite{SurveyDLCellularTraffic2023}. Recurrent-based architectures (e.g., ConvLSTMs) outperform classical models in timeseries prediction. ConvLSTM has been successfully applied to spatiotemporal forecasting tasks~\cite{ConvLSTM_NIPS15}. The xLSTM architecture~\cite{xlstm} was introduced to remedy key LSTM limitations such as the difficulty of updating long-term memories in the presence of newer or more relevant information and the restricted memory cell capacity. By reducing sequential dependencies in state updates, the xLSTM enables more efficient training on large datasets. 3D convolutional networks (Conv3D) have shown strong capabilities in learning spatiotemporal features~\cite{Zhang2023DST3D}. Graph-based approaches (e.g., Graph WaveNet) explicitly model dynamic spatial dependencies on graphs for traffic forecasting~\cite{Wu2019GraphWaveNet}. 

\begin{figure}
    \centering
    \includegraphics[width=\linewidth]{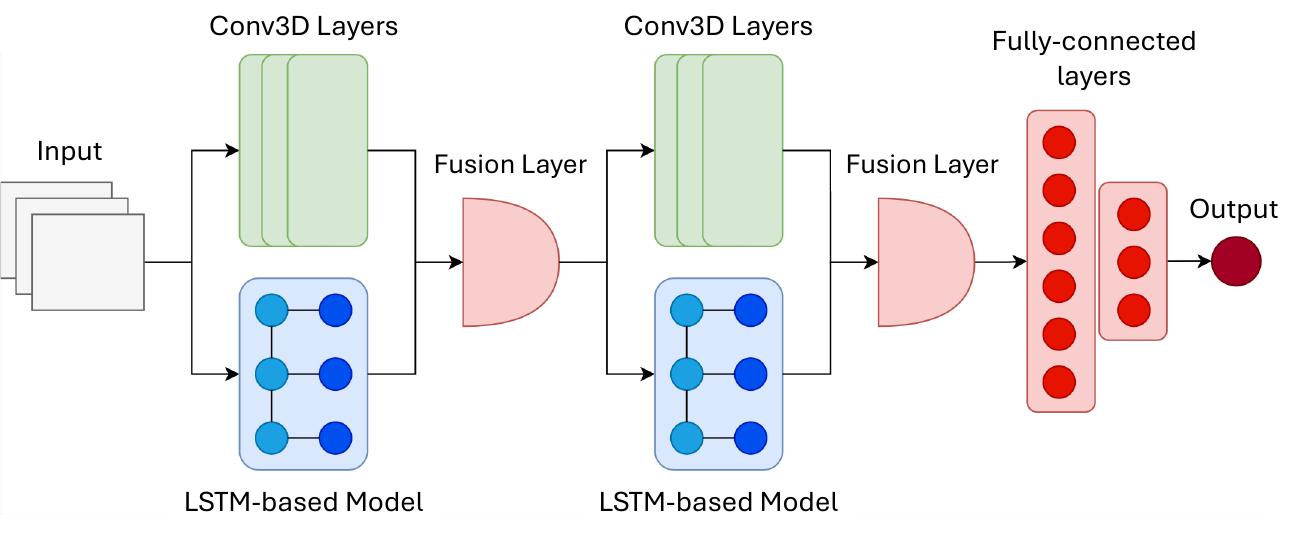}
    \caption{Architecture of the original STN, integrating ConvLSTM and Conv3D branches with a fusion and prediction module.}
    \label{fig:stn-arch}
\end{figure}

More recently, Transformer-based architectures tailored to traffic data have set new benchmarks by capturing long-range dependencies and regional heterogeneity in a unified framework, such as architectures presented in STGformer~\cite{STGormer2024}, STD-Net~\cite{STDNet2025}, and AMF-STGCN~\cite{Wang2021AMFSTGCN}. STN, a deep learning model, is designed for spatiotemporal mobile traffic forecasting~\cite{stn}. The STN architecture (Fig.~\ref{fig:stn-arch}) comprises two branches: a ConvLSTM-based temporal branch and a Conv3D-based spatial branch, capturing temporal and spatial correlations, respectively. The outputs are merged in a fusion layer to generate a unified feature space. The fused representation is passed to a prediction layer to generate the final output. However, the STN architecture exhibits several limitations:
\begin{itemize}
    \item \textbf{Computational Bottlenecks:} ConvLSTM models are computationally intensive, limiting their scalability.
    \item \textbf{Gradient Instability:} ConvLSTM exhibits inefficient gradient flow over long sequences, which inhibits model stability and slows convergence.
    \item \textbf{Rigid Fusion Strategy:} The fusion layer statically combines the output of the branches without adaptively emphasizing dominant spatial or temporal patterns. This limits its capacity to adjust to stochastic variations.
    \item \textbf{Generalization Constraints:} STN exhibits limited generalization in geographical regions, partly due to its reliance on hand-tuned fusion schemes and its lack of scalability to heterogeneous data distributions.
\end{itemize}

These limitations motivate the development of a more efficient and flexible spatiotemporal forecasting framework. In our approach, we introduce an enhanced STN that addresses these challenges through the integration of sLSTM units and cross-attention fusion for dynamic and adaptive feature integration.

\section{Methodology}\label{sec:methodology}
In this section, we formulate the problem of spatiotemporal forecasting and introduce our proposed architectures.

\subsection{Problem Formulation}
Spatiotemporal traffic forecasting can be modeled as a supervised learning problem where the input is the past $n$ observations (each observation is a \( I \times J \) spatial grid, see Fig.~\ref{fig:stn-arch}), and the labels are the observations at times $t + 1, \dots, t + \tau$, where $\tau$ is the forecast horizon. Unlike standard regression, this task involves predicting a variable based on its own lagged values, complicated by intricate dependencies among neighboring spatial timeseries. These latent interdependencies reduce prediction confidence as the forecast horizon increases, making long-term forecasting inherently more difficult. Spatiotemporal models address this challenge by capturing both spatial and temporal patterns. They aim to provide better accuracy than global models while avoiding the complexity of training a separate model for each cell.

We formally represent the network-wide mobile traffic forecasting problem over a time interval \( T \) as a spatiotemporal sequence of data points\footnote{A data point can represent the volume of mobile traffic uploads or downloads per time interval, the amount of SMS messages, etc.}:
\begin{equation}
\mathcal{X} = \{X_1, X_2, \ldots, X_T\},
\label{eq:1}
\end{equation}
where each \( X_t \) is a snapshot of the spatiotemporal timeseries at time \( t \), represented as a two-dimensional grid of size \( I \times J \):
\begin{equation}
X_t =
\begin{bmatrix}
x^{(1,1)}_t & \cdots & x^{(1,J)}_t \\
\vdots & \ddots & \vdots \\
x^{(I,1)}_t & \cdots & x^{(I,J)}_t
\end{bmatrix}.
\label{eq:2}
\end{equation}
Here, \( x^{(i,j)}_t \) denotes the series value located at coordinates \( (i, j) \) at time \( t \). The sequence \( \mathcal{X} \) can, thus, be interpreted as a tensor \( \mathcal{X} \in \mathbb{R}^{T \times I \times J} \). From an ML perspective, the spatiotemporal traffic forecasting problem involves predicting the most likely future sequence of \( \tau \) time steps \(\hat{X}_{t+1}, \hat{X}_{t+2}, \ldots, \hat{X}_{t+\tau}\) given the previous \( n \) observations. Formally, we aim to solve:
\begin{multline}
 \hat{X}_{t+1}, \ldots, \hat{X}_{t+\tau} = \\
 \operatorname*{argmax}_{X_{t+1}, \ldots, X_{t+\tau}} p(X_{t+1}, \ldots, X_{t+\tau} \mid X_{t-n+1}, \ldots, X_t).
\label{eq:3}   
\end{multline}

\begin{figure}
    \centering
    \includegraphics[width=1\linewidth]{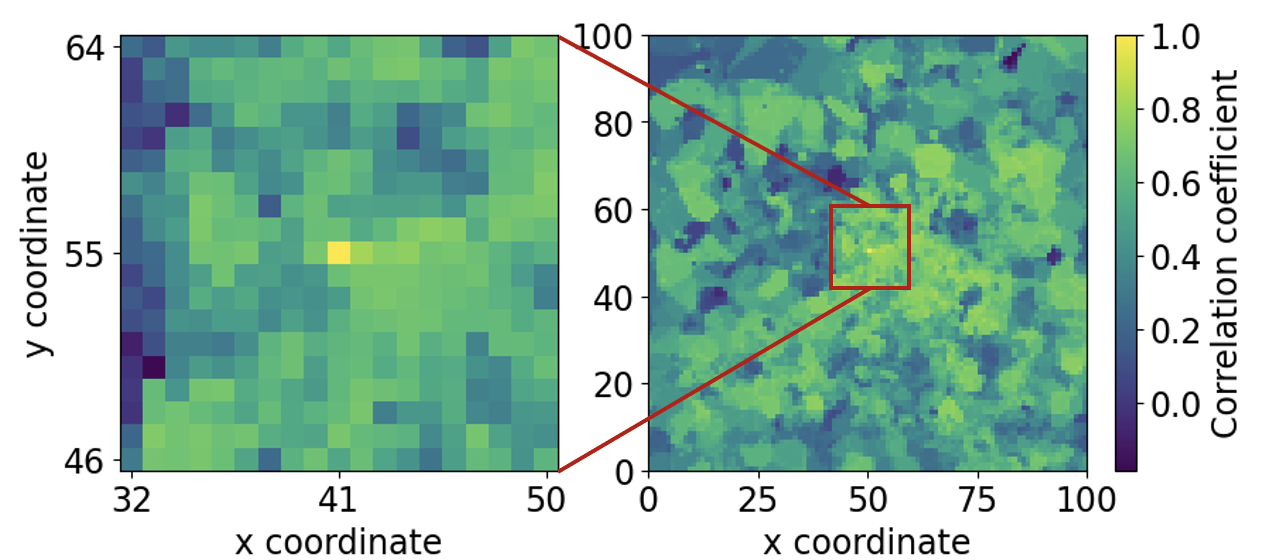}
    \caption{Spatial correlation between all the grid cells and the central one}
    \label{fig:correlation}
\end{figure}

Empirical studies suggest strong spatiotemporal correlations in mobile traffic patterns~\cite{Furno2017joint, wang2015understanding}. However, the traffic at the next time interval at any given cell \( x^{(i,j)}_{t+1} \) is influenced by traffic in its local neighborhood, while distant regions have minimal impact (see Fig.~\ref{fig:correlation})~\cite{li2014prediction}. Thus, the forecasting problem can be approximated by focusing on a local region of size \( (r+1) \times (r+1) \) surrounding each target cell.
Accordingly, the one-step forecasting distribution can be simplified as:
\begin{multline}
    p(X_{t+1} \mid X_{t-n+1}, \ldots, X_t) \approx \\ \prod_{i=1}^{I} \prod_{j=1}^{J} p\left(x^{(i,j)}_{t+1} \mid R^{(i,j)}_{t-n+1}, \ldots, R^{(i,j)}_t\right),
    \label{eq:4}
\end{multline}
where \( R^{(i,j)}_t \) denotes the local around cell \( (i,j) \) at time \( t \), defined as:
\begin{equation}
R^{(i,j)}_t =
\begin{bmatrix}
x^{(i - \frac{r}{2}, j - \frac{r}{2})}_t & \cdots & x^{(i + \frac{r}{2}, j - \frac{r}{2})}_t \\
\vdots & x^{(i,j)}_t & \vdots \\
x^{(i - \frac{r}{2}, j + \frac{r}{2})}_t & \cdots & x^{(i + \frac{r}{2}, j + \frac{r}{2})}_t
\end{bmatrix}.
\label{eq:5}
\end{equation}
The complete prediction over the entire grid at time \( t+1 \) is then represented as:
\begin{equation}
\hat{X}_{t+1} = \{ \hat{x}^{(i,j)}_{t+1} \mid i = 1, \ldots, I;\ j = 1, \ldots, J \}.
\label{eq:6}
\end{equation}
The value \( \hat{x}^{(i,j)}_{t+1} \) represents the prediction for the traffic volume in cell \( (i, j) \) at time \( t+1 \), and is obtained by solving:
\begin{equation}
\hat{x}^{(i,j)}_{t+1} = \operatorname*{argmax}_{x^{(i,j)}_{t+1}} p\left(x^{(i,j)}_{t+1} \mid R^{(i,j)}_{t-n+1}, \ldots, R^{(i,j)}_t\right).
\label{eq:7}
\end{equation}

These predictions depend on the marginal distribution of values within local neighborhoods, which the model must capture. The spatiotemporal problem is then reformulated as a learning mapping function \( \mathcal{F} \) that maps historical local data over fixed time steps to future data over fixed time steps:
\begin{multline}
    \hat{X}_{t+1}, \ldots, \hat{X}_{t+\tau} = \\ \Bigl\{\mathcal{F}_\theta \left( R^{(i,j)}_{t-n+1}, \ldots, R^{(i,j)}_t \right) \mid i = 1, \ldots, I;\ j = 1, \ldots, J\Bigr\}
    \label{eq:8}
  \end{multline}
where \( \theta \) represents the learnable parameters in the model.

\subsection{Spatiotemporal Network Architecture (STN)}
The original STN~\cite{stn} comprises two parallel branches: a temporal ConvLSTM-based branch and a spatial Conv3D branch, fused via a linear layer. Our enhanced STN replaces ConvLSTM with sLSTM for improved temporal modeling and introduces a transformer fusion layer instead of linear fusion.

\subsubsection{Temporal Branch} 
While the original STN used ConvLSTM, we adopt sLSTM from the xLSTM framework to overcome scalability and gradient issues. The forgetful aspect of LSTM for important information is improved in xLSTM via exponential gating and modified memory structures~\cite{xlstm}. The sLSTM from the xLSTM architecture provides a parameter-efficient alternative to conventional LSTMs with its scalar gating mechanism that stabilizes gradients and accelerates convergence. The gating mechanism is simplified by computing scalar gate values that are shared across the entire hidden state. The sLSTM improves state management with exponential gating for input and forget gates and multiple memory cells, enhancing its ability to store complex patterns.
\subsubsection{Spatial Branch} 
The spatial branch employs three Conv3D layers to extract hierarchical features from spatiotemporal grids~\cite{Medina2024CISTGCN}. Conv3Ds extend standard CNNs with temporal awareness and excel at modeling local variations, such as minor traffic fluctuations~\cite{Zhang2023DST3D,Kim20213DCNN,Vatamany2024GDCAF}. Given a sequence of spatiotemporal data with \(N\) feature maps \(X = \{X_1, X_2, \dots, X_N\}\), the spatial branch outputs feature maps \(H = \{H_1, \dots, H_M\}\) computed as:
\begin{equation}
h_m = act\left(\sum_{n=1}^N X_n \ast W_{mn} + b_m\right)
\label{eq:15}
\end{equation}
where \(\ast\) is 3D convolution and \(act(\cdot)\) is a nonlinear activation function. Unlike ConvLSTM, Conv3Ds do not backpropagate through time but maintain spatiotemporal locality via weight sharing, making them effective for short-term variation modeling and enhancing generalization.

\subsubsection{Fusion layer}
To replace the linear fusion in the original STN, we use a transformer-based layer to integrate spatial and temporal outputs and capture long-range dependencies~\cite{veit2016residual,Vaswani2017,Shaw2018,Chi2020}. This fusion leverages cross-attention, where spatial outputs \(\mathbf{X}_{\text{Spatial}}\) serve as queries and temporal outputs \(\mathbf{X}_{\text{Temporal}}\) as keys/values. The cross-attention scores are computed as:
\begin{multline}
  \mathrm{CrossAttention}(\mathbf{Q}_{\text{Spatial}}, \mathbf{K}_{\text{Temporal}}, \mathbf{V}_{\text{Temporal}}) = \\
  \mathrm{softmax}\left(\frac{\mathbf{Q}_{\text{Spatial}} \mathbf{K}_{\text{Temporal}}^\top}{\sqrt{d}}\right) \mathbf{V}_{\text{Temporal}},
  \label{eq:17}
\end{multline}
with \(d\) the embedding dimension~\cite{Vaswani2017}. Multi-head attention captures diverse feature relations; residual connections and layer normalization stabilize training~\cite{Ba2016,He2016}. The number of fusion blocks is a hyperparameter controlling the depth of this encoding.

 The fused features are decoded via an MLP to predict the marginal distribution
\(p(x^{(i,j)}_{t+1} \mid R^{(i,j)}_{t-n+1}, \ldots, R^{(i,j)}_t)\),
which estimates the traffic volume at location \((i,j)\) for time \(t+1\). The model \(\mathcal{M}\) parameterized by \(\theta\) maps input sequences \(\mathcal{R}^{(i,j)}_t = \{R^{(i,j)}_{t-n+1}, \ldots, R^{(i,j)}_t\}\) to predictions \(\hat{x}^{(i,j)}_{t+1}\):
\begin{equation}
  \hat{x}^{(i,j)}_{t+1} = \mathcal{M}(\theta; \mathcal{R}^{(i,j)}_t)
  \label{eq:21}
\end{equation}
trained by minimizing the L2 loss:
\begin{equation}
  L(\theta) = \frac{1}{T I J} \sum_{t=1}^T \sum_{i=1}^I \sum_{j=1}^J \|\mathcal{M}(\theta; \mathcal{R}^{(i,j)}_t) - x^{(i,j)}_t\|^2
  \label{eq:22}
\end{equation}

\section{Evaluation}\label{sec:eval}
In this section, we address three core research questions:
\begin{itemize}
    \item[\namedlabel{itm:first}{Q1})] \textbf{Predictive accuracy (Section~\ref{sec:accuracy}):} 
    (a) What do model behavior and prediction errors reveal? (\ref{sec:error_analysis}) 
    (b) How do architecture components affect forecast accuracy? (\ref{sec:accuracy_eval}, \ref{sec:gen}) 
    (c) How does our model compare to existing baselines? (\ref{sec:accuracy_eval})
    \item[\namedlabel{itm:second}{Q2})] \textbf{Robustness (Section~\ref{sec:gen}):} Does the model generalize to unseen spatial/temporal distributions?
    \item[\namedlabel{itm:third}{Q3})] \textbf{Usability (Section~\ref{sec:complex}):} What are the computational implications of deploying each model, in terms of training (\ref{sec:model_size}) and inference efficiency (\ref{sec:inference})?
\end{itemize}
To answer these questions, we formulate an experimental setup and define metrics that enable a comprehensive comparison of how well each architecture captures spatiotemporal dynamics.

\subsection{Baselines}
To asses the impact of each component of our model architecture, we implemented the following STN variants:
\begin{itemize}
    \item STN (baseline)~\cite{stn} – ConvLSTM (time) + Conv3D (space), fused via a linear layer. 
    \item STN-TF – ConvLSTM + Conv3D, fused via Transformer Fusion.
    \item STN-sLSTM – sLSTM (time) + Conv3D (space), fused via a linear layer.
    \item STN-sLSTM-TF – sLSTM + Conv3D, fused via Transformer Fusion (our proposed model).
\end{itemize}
We also include two purely temporal baselines—standard LSTM and xLSTM—by flattening each \(11\times11\) grid patch to a vector (i.e., no spatial modeling). This evaluates whether spatiotemporal architectures outperform global‐only predictors.

All variants share the same data pipeline, training splits, loss functions, optimizer, and regularization. Any performance differences thus stem solely from architectural changes.

\subsection{Experimental Setup}
\subsubsection{Evaluation Metrics}
We benchmark accuracy, robustness (across spatial, temporal, and traffic magnitude shifts), and spatiotemporal behavior (capturing mobility patterns). These evaluations ensure the model not only produces accurate forecasts but also generalizes effectively to new regions, periods, and varying traffic conditions while preserving realistic usage dynamics. Mean Absolute Error (\textit{MAE}) quantifies average absolute prediction error in the target units. Root Mean Squared Error (\textit{RMSE}) captures the magnitude of errors while remaining interpretable across scales. The Coefficient of Determination ($R^2$ Score) measures the proportion of variance explained by the model, with values close to 1 indicating strong performance. Structural Similarity Index (\textit{SSIM}) evaluates perceptual and structural similarity, which is critical for maintaining spatial coherence in predictions. All metrics are averaged over the full grid unless stated otherwise.

\subsubsection{Data Exploration and Preprocessing}
We use the Italia dataset~\cite{dataset}, containing two regions: Milan (10000 cells) and Trentino (11466 cells) with five features (SMS-in/out, Call-in/out, Internet activity) at 10 min intervals (Table~\ref{tab:dataset}). Training and testing occur on Milan’s traffic; Trentino serves as an unseen test for generalization.

\begin{table}
\caption{Statistical summary of Milan and Trentino grids, showing regional disparities.}\label{tab:dataset}
\centering
\begin{tabular}{l|r|r}
\hline
\bfseries  & \bfseries Trentino & \bfseries Milan\\
\hline
Number of samples & 8.928 & 8.928\\
Mean traffic value & 95.872 & 621.964\\
Standard Deviation & 37.812 & 223.385\\
Minimum & 39.934 &  20.811\\
Lower than 25\% & 65.270 & 413.368\\
Lower than 50\% & 98.718 &  64.178\\
Lower than 75\% & 111.219 & 810.414\\
Maximum & 395.858 & 1.234.958\\
\hline
\end{tabular}
\end{table}

\begin{figure}
\centerline{\includegraphics[width=\linewidth]{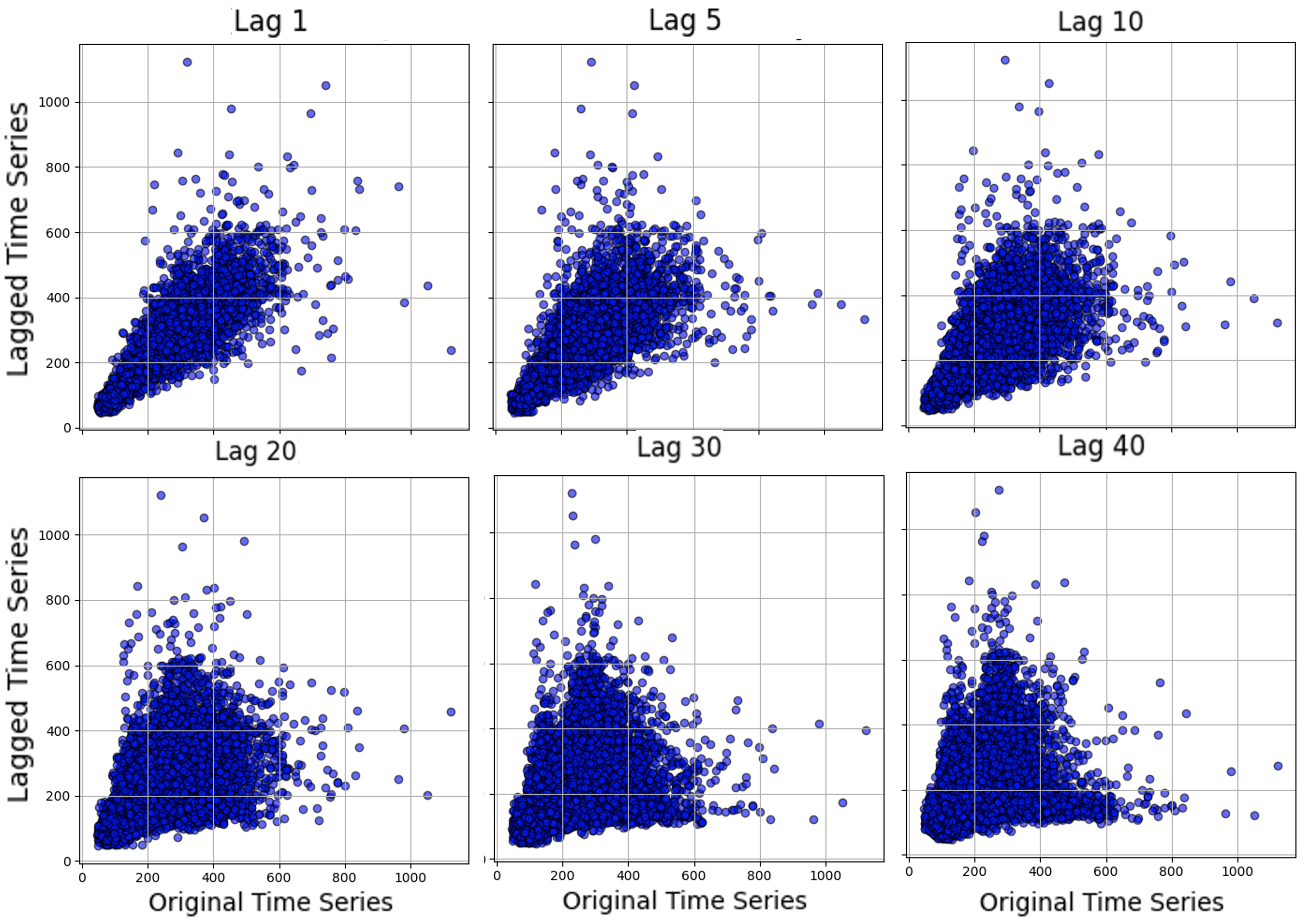}}
\caption{Autocorrelation Lag plots of a single cell for lag values 1 to 40.}\label{fig:single-scatter-autocorrelation-lags}
\end{figure}

At the single‐cell level, timeseries are highly variable with low autocorrelation (Approx. Entropy $\approx$1.386 for a random urban cell in Milan; Fig.~\ref{fig:single-scatter-autocorrelation-lags}), making univariate forecasting difficult. This is further exacerbated by spatial heterogeneity—higher activity in the center, clusters elsewhere (Fig.~\ref{fig:spatial_snapshot}). Adding neighbor cells captures marginal distributions (Section~\ref{sec:methodology}), enhancing cyclical patterns and reducing approximate entropy to 0.196. An Augmented Dickey–Fuller test confirms that, despite local non‐stationarity, the grid‐level series is stationary. By integrating spatial context, the model leverages cross-cell correlations to reinforce circadian and weekly cycles, effectively smoothing spikes. However, pronounced variability and low autocorrelation on a single-cell scale underscore the need for hybrid architectures that balance local adaptability with global structural coherence.

For preprocessing the dataset, we apply an \(11\times11\) kernel (\(r=5\)) to capture local spatial context, use six past time steps (\(n=6\)) to predict the next (\(\tau=1\)), and process edges via replication (‘edge’ padding), wherein input boundaries are extended by replicating edge values. Data are normalized per cell. Models train on 1 million samples (70\% of Milan, stride = 6) and test on 7 million samples (15\% of Milan, stride = 1).

\subsubsection{Experimental Setup}
We evaluate one‐step (10 min ahead) and multi‐step (6‐step autoregressive, 60 min ahead) forecasts. All models (PyTorch/PyTorch Lightning) run on a server with NVIDIA A100 (80 GB VRAM), 512 GB RAM, and AMD EPYC 75F3 CPUs.

Hyperparameter tuning for the STN-sLSTM-TF variant is performed with Optuna—this single optimization suffices since other variants (STN-sLSTM and STN-TF) share the same component hyperparameters. The search explores key architectural parameters (hidden size, number of STN blocks, sLSTM heads and layers, and fusion heads), and Table~\ref{tab:train_stats} reports the top five configurations from 24 trials. Baseline STN uses the optimal settings from~\cite{stn}. 
Models train for 40 epochs with Adam (\(\beta_1=0.9\), \(\beta_2=0.999\), \(\epsilon=10^{-8}\)) and initial LR \(5\times10^{-4}\), using batch normalization to speed convergence.~\cite{kingma2017adammethodstochasticoptimization}

\begin{table}
  \caption{Training Statistics and Model Complexity (5 Epochs)}\label{tab:train_stats}
  \centering
  \resizebox{\columnwidth}{!}{%
  \begin{tabular}{l|rrrrc}
    \hline
    \textbf{Config} & \textbf{Train Loss} & \textbf{Val Loss} & \makecell{\textbf{Best}\\\textbf{Loss}} & \makecell{\textbf{Gap}\\\textbf{(Ovr)}}  & \textbf{Params (K)} \\
    \hline
    \makecell{\textbf{(h=64,b=2,}\\\textbf{a=4,l=2,f=8)}} & \makecell{0.179612\\$\pm$0.000293} & \makecell{0.141557\\$\pm$0.000003} & 0.141555& 0.038055 & 156.2 \\
    \hline
    \makecell{(h=64,b=2,\\a=2,l=1,f=8)} & \makecell{0.179612\\$\pm$0.000036} & \makecell{0.141557\\$\pm$0.000003} & 0.141555 & 0.038055 & 127.1 \\
    \hline
    \makecell{(h=64,b=1,\\a=8,l=1,f=2)} & \makecell{0.179612\\$\pm$0.000025} & \makecell{0.141557\\$\pm$0.000003} & 0.141555 & 0.038055 & 152.3 \\
    \hline
    \makecell{(h=64,b=2,\\a=8,l=2,f=2)} & \makecell{0.17366\\$\pm$0.001592} & \makecell{0.147424\\$\pm$0.000459} & 0.146774 & 0.026237 & 245.2 \\
    \hline
    \makecell{(h=64,b=2,\\a=8,l=1,f=8)} & \makecell{0.173744\\$\pm$0.001339} & \makecell{0.147114\\$\pm$0.00044} & 0.146604 & 0.026629 & 183.5 \\
    \hline
  \end{tabular}%
  }

    \vspace{1ex}

  \raggedright\textit{Key:}
  h = hidden size, 
  b = \# STN blocks, 
  a = \# sLSTM heads, 
  l = \# sLSTM layers, 
  f = \# fusion Transformer heads.
  \vspace{1ex}

  \raggedright\textit{Note:}  
  Train/Val/Best = mean$\pm$std loss over epochs;  
  Gap = Val-Best (overfitting indicator);  
  Params in thousands.

\end{table}

\subsection{Prediction accuracy}\label{sec:accuracy}
To answer Question~\ref{itm:first}, we conduct error analysis to validate the accuracy. We also perform an ablation study (STN vs. STN-TF vs. STN-sLSTM vs. STN-sLSTM-TF).

\subsubsection{Error Analysis} \label{sec:error_analysis}
We retrained the optimal STN-sLSTM-TF (hidden size 64, 2 STN blocks, 4 sLSTM heads, 2 sLSTM layers, and 8 fusion heads) and evaluated on the test set. 

\begin{figure}
  \centering
  \includegraphics[width=\linewidth]{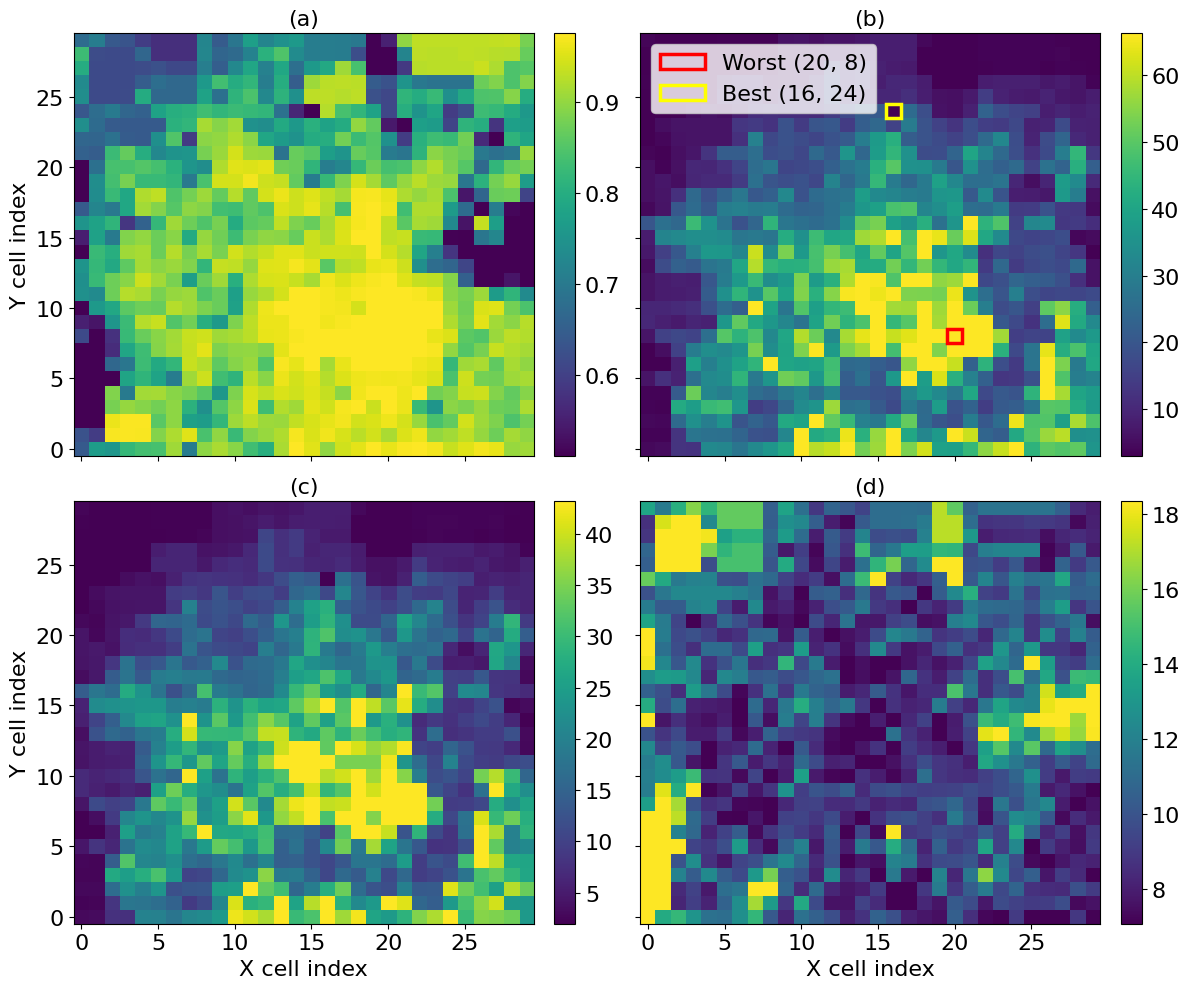}
  \caption{Spatial heatmap of (a) R², (b) RMSE, (c) MAE, and (d) MAPE on the test set.}
  \label{fig:r2_heatmap}
\end{figure}

Fig.~\ref{fig:r2_heatmap} shows spatial heatmaps for $R^2$, RMSE, MAE, and MAPE, where each pixel represents the metric value for the timeseries in a given spatial cell. Central, high‐traffic cells achieve $R^2>0.90$, while peripheral scores lie between 0.75–0.85. Similar spatial trends appear in the remaining metrics. The best and worst cells (Fig. \ref{fig:r2_heatmap}b) suggest that region‐aware embeddings or geographically informed regularization could improve variable zones.

\begin{figure}
  \centering
  \includegraphics[width=\linewidth]{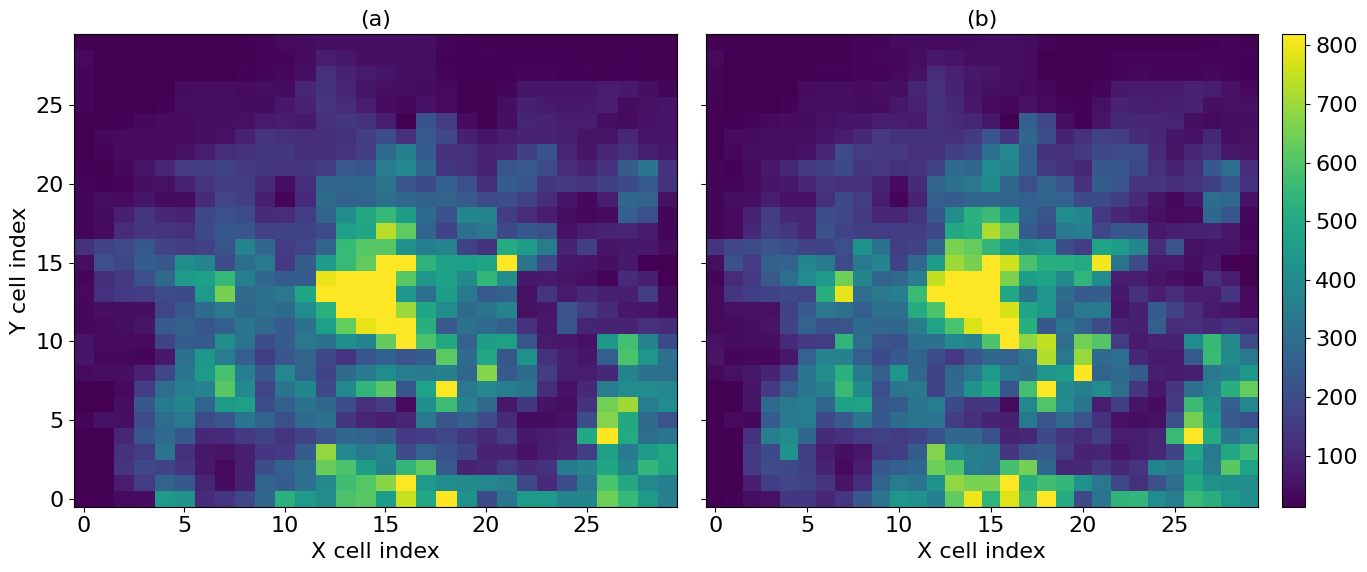}
  \caption{Full-grid spatial heatmaps of (a) actual vs.\ (b)  predicted values at 2013-12-08 @00:10, showing preservation of overall spatial patterns.}
  \label{fig:spatial_snapshot}
\end{figure}

Fig.~\ref{fig:spatial_snapshot} shows a single‐timestamp full‐grid prediction, generated by predicting each cell to reconstruct the full-grid. The Figure confirms that macro-level spatial gradients are well preserved, but low-density residential areas are slightly underestimated, indicate potential gains from integrating auxiliary covariates (e.g., land use, weather) to correct residual errors.

\begin{figure}
  \centering
  \includegraphics[width=\linewidth]{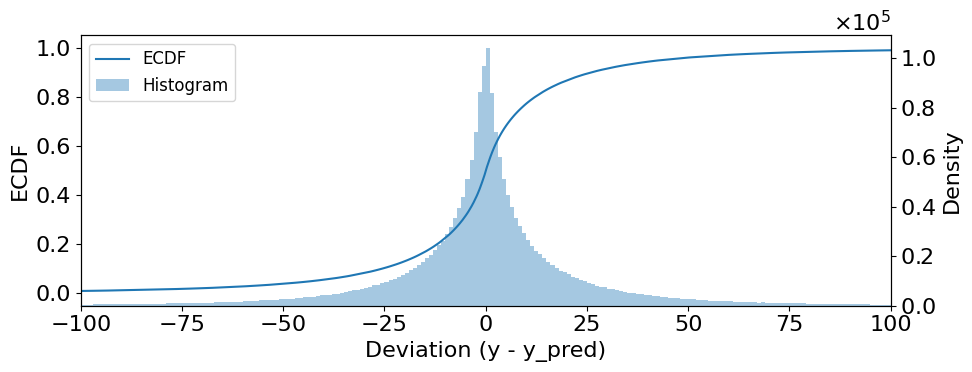}
  \caption{ECDF and histogram of deviations \((y - \hat y)\).}\label{fig:error_histogram}
\end{figure}

\begin{figure}
  \centering
  \includegraphics[width=\linewidth]{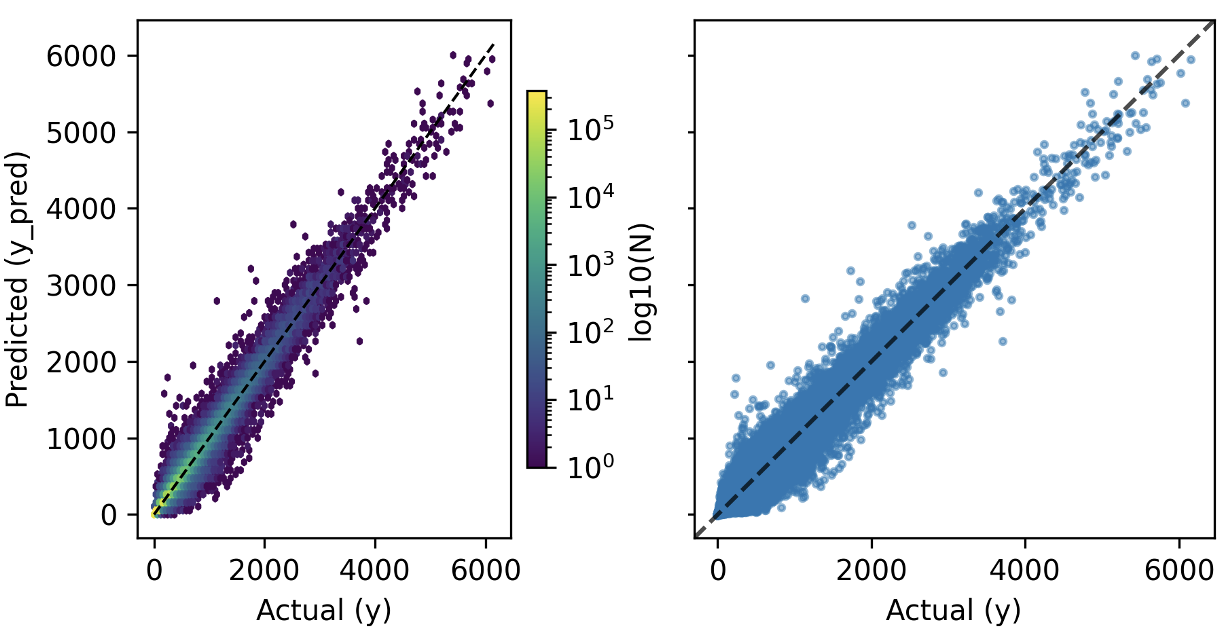}
  \caption{Scatter plot of predicted vs.\ actual values; deviations from the diagonal highlight mispredictions.}
  \label{fig:scatter_pred_actual}
\end{figure}

Fig.~\ref{fig:error_histogram} shows the error histogram and empirical cumulative distribution function (ECDF) of single‐step deviations across Milan grid. The histogram is slightly right‐skewed (skew \(\approx 0.3\)), implying underprediction during peaks—an asymmetric loss or bias correction may help. The ECDF shows that 80\% of the deviation lies within \(\pm1.2\) units, defining tight overall error bounds. However, 2.08\% of the values exhibit deviations exceeding \(\pm100\) units (max 1439.70, min –1685.59). These outliers highlight the need for targeted handling strategies.
Fig.~\ref{fig:scatter_pred_actual} plots predicted vs.\ actual values: points cluster tightly along \(y=x\) (Pearson \(r\approx0.98\)), with isolated extremes suggesting potential benefits from a tail‐focused ensemble.

\begin{figure}
  \centering
  \includegraphics[width=\linewidth]{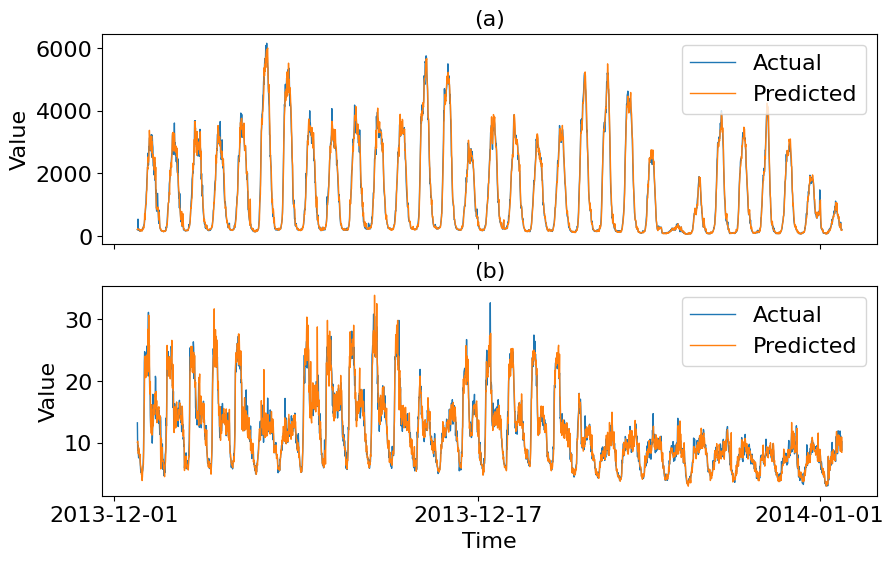}
  \caption{Timeseries of actual vs.\ predicted values for (a) worst- and (b) best-performing cells in term of RMSE, highlighting trend fidelity and peak alignment (cf. Fig.~\ref{fig:r2_heatmap}b)}
  \label{fig:time_series_plot}
\end{figure}

To evaluate performance over time, we selected the best cell (16, 24) and worst cell (20, 8) by RMSE (Fig. \ref{fig:r2_heatmap}b) and plotted December 2013’s actual vs.\ predicted traffic (Fig. \ref{fig:time_series_plot}). Fig.~\ref{fig:time_series_plot}a shows the predicted traffic volume for the worst-performing and Fig.\ref{fig:time_series_plot}b shows the best-performing cell, suggesting that the model captures both seasonal trends and intra-hour fluctuations. 
Performance differences between the two cells can be partially attributed to traffic magnitude and variability. The worst-performing cell, in a high-traffic city center, sees abrupt spikes—especially during events—making prediction harder. In contrast, the best-performing rural cell has low, stable traffic, easing forecasting. These magnitude differences inflate RMSE and MAE, revealing a broader challenge: models struggle to balance accuracy across regions with varying scales and dynamics. Large-magnitude series dominate the loss, often degrading performance on smaller ones. Addressing this requires techniques that can account for scale heterogeneity and variance in temporal dynamics.

\begin{figure*}
  \centering
  \includegraphics[width=\textwidth]{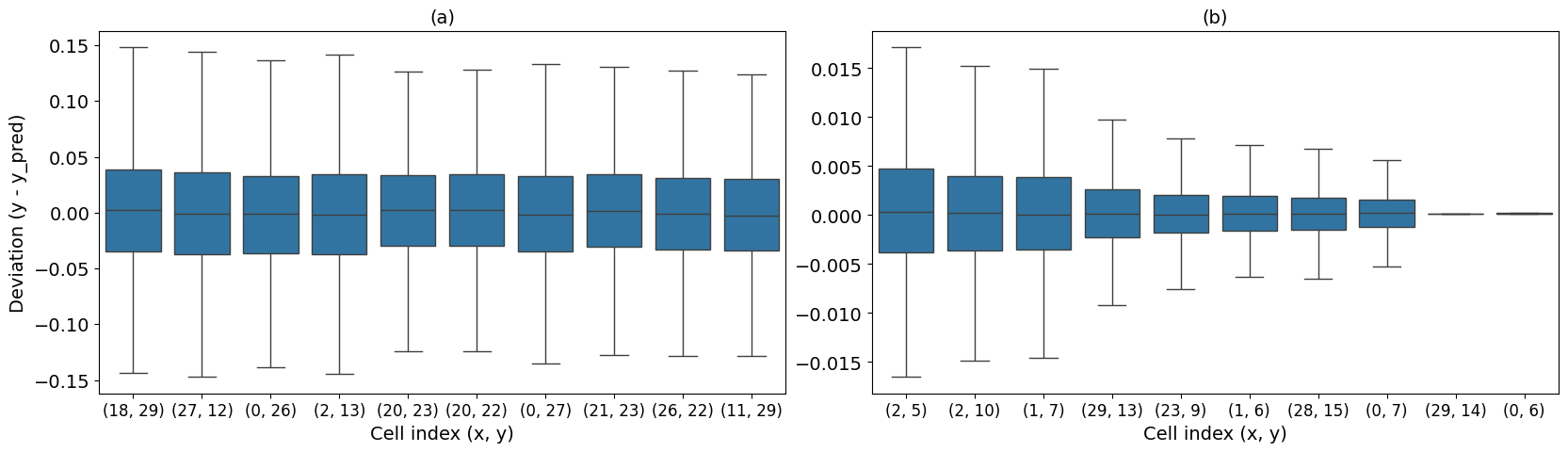}
  \caption{Boxplots of deviations \((y - \hat{y})\) for the (a) worst- and (b) best-performing cells in terms of MAE, showing consistency vs.\ variability in error.}
  \label{fig:cluster_boxplot}
\end{figure*}

Finally, the boxplots in Fig.\ref{fig:cluster_boxplot} compare the normalized deviations for bottom‐ (Fig.\ref{fig:cluster_boxplot}a) and top‐MAE (Fig.\ref{fig:cluster_boxplot}b) cell clusters. The top-performing cells exhibit tightly centered deviations with an interquartile range (IQR) of approximately 0.01, indicating high predictive accuracy. In contrast, the worst-performing cells show larger spread and bias, with IQRs $\approx$ 0.1 and positive skew (implying underprediction in high-magnitude cells). This discrepancy emphasizes the importance of adaptive loss weighting or cluster-aware calibration to mitigate localized prediction errors in heterogeneous areas.

\textit{Overall, the model delivers strong spatial and temporal fidelity; however, heterogeneity in both magnitude and spatial dynamics introduces localized prediction challenges that are not fully addressed by global optimization alone.} These observations point to promising future directions, including spatially adaptive modeling strategies, scale-aware loss functions, and the incorporation of auxiliary contextual features to enhance generalization and reduce localized error spikes.

\subsubsection{Performance Evaluation} \label{sec:accuracy_eval}
Table~\ref{tab:model_performance} compares STN-sLSTM-TF with baselines. As can be seen, our proposed approach achieves the lowest error metrics yielding the best forecast performance across almost all metrics.
Replacing ConvLSTM with sLSTM yields consistent improvements across all metrics, highlighting sLSTM's effectiveness in capturing long-term temporal dependencies. Although xLSTM outperforms STN in the Milan dataset, it shows weaker generalization. Incorporating Transformer-based fusion (models with the suffix -TF) provides modest but consistent gains for both STN and STN-sLSTM, improving spatial feature integration.

\begin{table}
    \centering
    \caption{Performance comparison for single time-step forecast on Milan dataset. \textbf{Bold} indicates the best performance, and \underline{underline} indicates the second best.}\label{tab:model_performance}
    \begin{tabular}{l|cccc}
    \hline
    \textbf{Variant} & \textbf{MAE $\downarrow$} & \textbf{RMSE $\downarrow$}& \textbf{R² Score $\uparrow$ }& \textbf{SSIM $\uparrow$} \\
    \hline
    LSTM & 101.468 & 157.91 & 0.3170 & 0.6068 \\
    xLSTM & 6.4672 & 15.0901 & 0.9637 & 0.9870 \\
    STN      & 7.3917 & 16.8849 & 0.9546 & 0.9853 \\
    STN-TF   & 7.2104 & 15.5359 & 0.9616 & 0.9858 \\
    STN-sLSTM   & \underline{5.6319} & \underline{12.5036} & \underline{0.9751} & \underline{0.9912} \\
    STN-sLSTM-TF     & \textbf{5.5375} & \textbf{12.0271} & \textbf{0.9770} & \textbf{0.9921} \\
    \hline
    \end{tabular}
\end{table}

In order to evaluate the forecast performance multiple timesteps ahead, we implemented  autoregressive forecasting using six iterative steps (60 minute ahead, using 60 minute input data points). As shown in Table~\ref{tab:performance_comparison}, STN-sLSTM consistently achieves lower errors and higher SSIM across all steps, while STN-sLSTM-TF shows the best RMSE and $R^2$ with just marginally worse MAE compared to STN-sLSTM. STN's performance degrades over time due to ConvLSTM’s limited capacity for capturing long-term dependencies. In contrast, STN-sLSTM maintains stable predictions and spatial coherence over longer time horizons. Although STN-sLSTM yields the lowest MAE, STN-sLSTM-TF performs better in terms of RMSE across the full sequence, suggesting superior ability to track peaks in high-magnitude regions. This performance gain can be partially attributed to the transformer’s enhanced capacity for fusing spatial feature representations, particularly in high-traffic areas. These results underscore the benefits of attention-based fusion mechanisms in improving long-range forecasting accuracy and maintaining structural consistency.

\begin{table}
    \centering
    \caption{Autoregressive performance comparison (next-60-minutes prediction in the Milan dataset). \textbf{Bold} indicates the best performance, and \underline{underline} indicates the second best.}\label{tab:performance_comparison}
    \begin{tabular}{cl|ccc}
        \hline
        \textbf{Step} & \textbf{Variant}   & \textbf{MAE $\downarrow$} & \textbf{RMSE $\downarrow$}& \textbf{SSIM $\uparrow$} \\
        \hline
        \multirow{3}{*}{1} 
            & STN & 4.3109 & 10.3451 & 0.9726 \\
            & STN-sLSTM & \textbf{3.3553} & \underline{8.2769} & \textbf{0.9813} \\
            & STN-sLSTM-TF & \underline{3.4967} & \textbf{7.9149} & \underline{0.9798} \\
        \hline
        \multirow{3}{*}{2} 
            & STN & 5.2234 & 12.7886 & 0.9604 \\
            & STN-sLSTM & \textbf{3.8980} & \underline{9.8735} & \textbf{0.9755} \\
            & STN-sLSTM-TF & \underline{4.0505} & \textbf{9.0952} & \underline{0.9742} \\
        \hline
        \multirow{3}{*}{3} 
            & STN & 5.9896 & 14.4512 & 0.9495 \\
            & STN-sLSTM & \textbf{4.3887} & \underline{10.8846} & \textbf{0.9706} \\
            & STN-sLSTM-TF & \underline{4.5854} & \textbf{9.8406} & \underline{0.9691} \\
        \hline
        \multirow{3}{*}{4} 
            & STN & 7.1585 & 16.8735 & 0.9319 \\
            & STN-sLSTM & \textbf{5.3422} & \underline{12.7931} & \textbf{0.9606} \\
            & STN-sLSTM-TF & \underline{5.5740} & \textbf{11.5626} & \underline{0.9590} \\
        \hline
        \multirow{3}{*}{5} 
            & STN & 8.0777 & 18.9370 & 0.9185 \\
            & STN-sLSTM & \textbf{6.0010} & \underline{14.3070} & \textbf{0.9543} \\
            & STN-sLSTM-TF & \underline{6.3459} & \textbf{12.9664} & \underline{0.9521} \\
        \hline
        \multirow{3}{*}{6} 
            & STN & 9.0152 & 20.5406 & 0.9018 \\
            & STN-sLSTM & \textbf{6.6677} & \underline{15.2810} & \textbf{0.9468} \\
            & STN-sLSTM-TF & \underline{7.1529} & \textbf{14.0592} & \underline{0.9432} \\
        \hline
    \end{tabular}
\end{table}

\subsection{Model Generalization}\label{sec:gen}
To address Question~\ref{itm:second} on robustness, we evaluated models trained on Milan directly on the unseen Trentino dataset to assess generalization capabilities. As shown in Table~\ref{tab:trentino_performance}, STN-sLSTM and STN-sLSTM-TF outperform all baselines, achieving lower MAE and RMSE and higher SSIM. STN-sLSTM attains the lowest MAE (1.6803) and strong RMSE (5.3016), surpassing both the original STN and xLSTM. While STN-TF shows modest improvements across metrics, it remains behind the sLSTM variants. High SSIM scores (0.9903 for STN-sLSTM, 0.9884 for STN-sLSTM-TF) confirm their ability to preserve spatial structure on new distributions. In the autoregressive task (see Table~\ref{tab:autoregressive_performance}), STN's performance degrades rapidly, with SSIM dropping below 0.94 by the sixth step. STN-sLSTM maintains stable predictions, with MAE under 2.0 and SSIM above 0.96 throughout. STN-sLSTM-TF performs second-best, particularly in mid-range steps, suggesting that transformer fusion enhances resilience to distributional shifts by complementing recurrent modeling.

\begin{table}
    \centering
    \caption{Performance comparison for single time-step forecast on an unseen dataset (Trentino). \textbf{Bold} indicates the best performance, and \underline{underline} indicates the second best.}\label{tab:trentino_performance}
    \begin{tabular}{l|cccc}
        \hline
        \textbf{Variant} & \textbf{MAE $\downarrow$} & \textbf{RMSE $\downarrow$}& \textbf{R² Score $\uparrow$ }& \textbf{SSIM $\uparrow$} \\
        \hline
        xLSTM & 2.5974 & 8.9235 & 0.8793 & 0.9615 \\
        STN & 2.6344  & 7.6370  & 0.9116  & 0.9744  \\
        STN-TF & 2.6059  & 6.8237  & 0.9294  & 0.9804  \\
        STN-sLSTM & \textbf{1.6803}  & \underline{5.3016}  & \underline{0.9574}  & \textbf{0.9903}  \\
        STN-sLSTM-TF & \underline{2.1257} & \textbf{5.2649}  & \textbf{0.9580}  & \underline{0.9884}  \\
        \hline
    \end{tabular}
\end{table}

\begin{table}
    \centering
    \caption{Autoregressive performance comparison (next-60-minutes prediction on unseen Trentino dataset). \textbf{Bold} indicates the best performance, and \underline{underline} indicates the second best.}\label{tab:autoregressive_performance}
    \begin{tabular}{cl|ccc}
        \hline
        \textbf{Step} & \textbf{Variant}   & \textbf{MAE $\downarrow$} & \textbf{RMSE $\downarrow$}& \textbf{SSIM $\uparrow$} \\
        \hline
        \multirow{3}{*}{1}
            & STN               & \underline{1.3025} & 5.4483 & \underline{0.9852} \\
            & STN-sLSTM                  & \textbf{0.8662}    & \textbf{2.6973}    & \textbf{0.9925}    \\
            & STN-sLSTM-TF    & 1.4498             & \underline{3.3183}             & 0.9726             \\
        \hline
        \multirow{3}{*}{2}
            & STN               & \underline{1.5540} & 6.8674             & \underline{0.9763} \\
            & STN-sLSTM                  & \textbf{1.0356}    & \textbf{3.3212}    & \textbf{0.9889}    \\
            & STN-sLSTM-TF    & 1.5867             & \underline{3.7074} & 0.9687             \\
        \hline
        \multirow{3}{*}{3}
            & STN               & 1.7955             & 8.3007             & \underline{0.9677} \\
            & STN-sLSTM                  & \textbf{1.2108}    & \textbf{3.9695}    & \textbf{0.9852}    \\
            & STN-sLSTM-TF    & \underline{1.7614} & \underline{4.2253} & 0.9642             \\
        \hline
        \multirow{3}{*}{4}
            & STN               & 2.0917             & 9.3477             & \underline{0.9550} \\
            & STN-sLSTM                  & \textbf{1.5267}    & \textbf{4.7829}    & \textbf{0.9768}    \\
            & STN-sLSTM-TF    & \underline{2.0396} & \underline{5.0625} & 0.9548             \\
        \hline
        \multirow{3}{*}{5}
            & STN               & 2.3192             & 10.3474            & 0.9447             \\
            & STN-sLSTM                  & \textbf{1.7694}    & \textbf{5.6378}    & \textbf{0.9693}    \\
            & STN-sLSTM-TF    & \underline{2.2539} & \underline{5.8353} & \underline{0.9483} \\
        \hline
        \multirow{3}{*}{6}
            & STN               & 2.5198             & 11.2963            & 0.9361             \\
            & STN-sLSTM                  & \textbf{1.9796}    & \textbf{6.4522}    & \textbf{0.9630}    \\
            & STN-sLSTM-TF    & \underline{2.4561} & \underline{6.5270} & \underline{0.9424} \\
        \hline
    \end{tabular}
\end{table}

\textit{The strong performance of STN-sLSTM variants on the Trentino dataset highlights their robust transferability to unseen domains.} Unlike STN, which fails to adapt its temporal and spatial filters effectively, sLSTM's flexible architecture generalizes well without retraining. Transformer-based fusion further enhances generalization at intermediate horizons by emphasizing key spatial features. These results suggest that future spatiotemporal models should prioritize recurrent units with improved memory retention, such as sLSTM, while using attention mechanisms as complementary components for enhanced adaptability across diverse settings.

\subsection{Computational Efficiency Analysis}\label{sec:complex}
To address~\ref{itm:third}, we perform experiments to assess computational complexity in terms of resource footprint and operational efficiency (inference time and MAC operations).

\subsubsection{Model Size and Memory Footprint}\label{sec:model_size}
Table~\ref{tab:model-comparison} shows that the STN model is very lightweight, requiring only 2.20 MB of storage and 11.34 MB of GPU memory. In comparison, STN-sLSTM variants occupy 6.22 MB on disk, with GPU memory usage increasing to 17.67 MB (STN-sLSTM) and 26.95 MB (STN-sLSTM-TF). This increase reflects the additional hidden-state storage and transformation parameters. STN-sLSTM and STN-sLSTM-TF use approximately 140\% and 238\% more GPU memory than STN, respectively.

\begin{table}
\centering
\caption{Computational efficiency comparison, showing resource footprint and computational complexity}\label{tab:model-comparison}
\resizebox{\columnwidth}{!}{%
\begin{tabular}{lcccc>{\centering\arraybackslash}p{1.5cm}>{\centering\arraybackslash}p{1.5cm}}
\toprule
Model & Size (MB) & \makecell{GPU\\Memory (MB)} & MACs & \multicolumn{2}{c}{Inference (ms)} \\
\cmidrule(lr){5-6}
& & & & \makecell{Cell} & \makecell{Grid} \\
\midrule
STN (baseline) & 2.20 & 11.34 & $2.31 \times 10^6$ & 2.58 & 28.23 \\
STN-sLSTM & 6.22 & 17.67 & $1.02 \times 10^7$ & 7.66 & 34.17 \\
STN-sLSTM-TF & 6.22 & 26.95 & $1.02 \times 10^7$ & 8.08 & 326.01 \\
\bottomrule
\end{tabular}%
}
\end{table}

\subsubsection{Inference Speed and Computational Complexity}\label{sec:inference}
STN performs $2.31 \times 10^6$ multiply-accumulate operations (MACs), significantly lower than the $1.02\times10^7$ MACs required by both STN-sLSTM variants. This 4x complexity increase  manifests in inference time: STN processes a single cell in 2.58 ms versus 7.66 ms (STN-sLSTM) and 8.08 ms (STN-sLSTM-TF). At grid level, latency rises sharply to 326.01 ms for STN-sLSTM-TF due to transformer fusion overhead. Crucially, cell-level inference remains below 10 ms across all variants, well within near-real-time requirements ($<$100ms) for control loops. While grid-level latency is substantial, we note this can be mitigated through horizontal scaling, and future work could apply pruning/quantization techniques for further optimization.

\subsubsection{Trade-offs and Practical Implications}
Our benchmarks reveal a fundamental efficiency-capacity trade-off: \textit{STN excels in low-resource scenarios, while STN-sLSTM variants deliver stronger temporal modeling at higher computational cost.} STN-sLSTM without fusion balances these priorities for moderate-complexity deployments. Although STN-sLSTM-TF incurs significant grid-level latency, it provides the most stable and accurate forecasts in high-magnitude, high-variability areas; particularly valuable when dataset heterogeneity demands sophisticated spatiotemporal fusion. For real-time control applications, cell-level inferences enable immediate actions and outweigh grid-level limitations.

\section{Conclusion and Future Work}\label{conclusion}
We introduced an sLSTM-enhanced STN for 5G mobile traffic forecasting, combining a temporal sLSTM branch and a three-layer Conv3D path for spatial feature extraction, fused via a Transformer layer. This design improved gradient stability over ConvLSTM and strengthened feature integration. Empirical results showed a 23\% MAE reduction over the original STN and a 30\% improvement on unseen data, highlighting strong generalization. While Transformer fusion boosted long-range interaction modeling, its complexity may challenge scalability in resource-limited contexts. Our STN variants outperformed baselines in both static and autoregressive settings, with the Transformer proving especially beneficial for long-term forecasting due to its handling of intricate spatiotemporal dependencies. These results advocate for sLSTM-based architectures with advanced fusion mechanisms.

Future work will focus on three key directions: First, we will conduct comprehensive benchmarking against contemporary spatiotemporal models including graph-based approaches, transformer architectures, and emerging diffusion/transformer hybrids to establish relative performance boundaries. Second, we plan deeper analysis of the transformer fusion mechanism to elucidate how spatial and temporal representations interact during cross-attention. Third, we will validate generalization capabilities across diverse spatiotemporal datasets with varying statistical properties to stress-test domain adaptability. Complementing these, we will explore integrating mLSTM with Conv3D for enhanced pattern capture, cascading mLSTM with sLSTM into unified xLSTM structures, and replacing Conv3D with mLSTM for joint spatiotemporal modeling to address current integration limitations while enhancing flexibility.

\section*{Acknowledgement}
This work was partly funded by the Bavarian Government by the Ministry of Science and Art through the HighTech Agenda (HTA).

\end{document}